\documentclass[conference]{IEEEtran}
\IEEEoverridecommandlockouts
\usepackage{cite}
\usepackage{amsmath,amssymb,amsfonts}
\usepackage{algorithmic}
\usepackage{graphicx}
\usepackage{textcomp}
\usepackage{xcolor}
\usepackage{booktabs}
\usepackage{multirow}
\def\BibTeX{{\rm B\kern-.05em{\sc i\kern-.025em b}\kern-.08em
    T\kern-.1667em\lower.7ex\hbox{E}\kern-.125emX}}
\begin{document}

\title{MAFNet: Multi-frequency Adaptive Fusion Network for Real-time Stereo Matching}

\author{

\IEEEauthorblockN{
Ao~Xu\textsuperscript{2},
Rujin~Zhao\textsuperscript{3,6},
Xiong~Xu\textsuperscript{1,2,4,5}\thanks{Corresponding author: xvxiong@tongji.edu.cn},
Boceng~Huang\textsuperscript{1},
Yujia~Jia\textsuperscript{2}
}

\IEEEauthorblockN{
Hongfeng~Long\textsuperscript{3},
Fuxuan~Chen\textsuperscript{1},
Zilong~Cao\textsuperscript{1},
Fangyuan~Chen
}

\IEEEauthorblockA{
\textsuperscript{1}College of Surveying and Geo-Informatics, Tongji University, Shanghai, China\\
\textsuperscript{2}Shanghai Research Institute for Intelligent Autonomous Systems, Tongji University, Shanghai, China\\
\textsuperscript{3}Institute of Optics and Electronics, Chinese Academy of Sciences, Chengdu 610209, China\\
\textsuperscript{4}Shanghai Integrated Innovation Center for Manned Lunar Exploration, Shanghai, China\\
\textsuperscript{5}Shanghai Key Laboratory for Planetary Mapping and Remote Sensing for Deep Space Exploration, Shanghai, China\\
\textsuperscript{6}University of Chinese Academy of Sciences, Beijing 100049, China
}

}

\maketitle

\begin{abstract}
Existing stereo matching networks typically rely on either cost-volume construction based on 3D convolutions or deformation methods based on iterative optimization. The former incurs significant computational overhead during cost aggregation, whereas the latter often lacks the ability to model non-local contextual information. These methods exhibit poor compatibility on resource-constrained mobile devices, limiting their deployment in real-time applications. To address this, we propose a Multi-frequency Adaptive Fusion Network (MAFNet), which can produce high-quality disparity maps using only efficient 2D convolutions. Specifically, we design an adaptive frequency-domain filtering attention module that decomposes the full cost volume into high-frequency and low-frequency volumes. Subsequently, we introduce a Linformer-based low-rank attention to adaptively aggregation high- and low-frequency information, yielding more robust disparity estimation. Extensive experiments demonstrate that the proposed MAFNet significantly outperforms existing real-time methods on public datasets such as Scene Flow and KITTI 2015, showing a favorable balance between accuracy and real-time performance. Code:https://anonymous.4open.science/r/MAFNet-B05E.
\end{abstract}


\section{Introduction}
\label{sec:intro}

With the advent of autonomous driving, drone inspection, smart manufacturing, augmented reality, and medical robotics, acquiring accurate 3D structural information has become essential for intelligent perception systems \cite{lu2025fast, zhang2023uavstereo, zhang2022deep, cheng2024stereo, xia2022robust}. Stereo matching, which estimates scene depth by establishing pixel correspondences between rectified image pairs under geometric constraints, is a key component of 3D vision \cite{liu2024playing}. Leveraging deep learning, end-to-end stereo networks \cite{xu2023iterative, guan2025bridgedepth, wen2025foundationstereo, cheng2025monster} have achieved state-of-the-art results on benchmarks such as Scene Flow \cite{mayer2016large} and KITTI \cite{menze2015object}, driving the adoption of 3D vision across research and industry.

Recent studies have extensively explored cost-volume construction and representation to improve stereo matching \cite{zeng2023parameterized, xu2025igev++}. Representative methods include GwcNet \cite{guo2019group}, which computes group-wise correlations to form more discriminative volumes; ACVNet \cite{xu2022attention}, which introduces attention mechanisms to convert correlation volumes into attention weights for finer filtering; and PCWNet \cite{shen2022pcw}, which combines pyramidal structures with deformable volumes for enhanced multi-scale modeling. While these approaches significantly boost accuracy, their reliance on high-resolution volumes and large-scale 3D convolutions incurs heavy computational and memory costs, hindering real-time deployment on resource-constrained mobile or embedded platforms.

\begin{figure}[htbp]
\centering
\includegraphics[width=0.9\columnwidth]{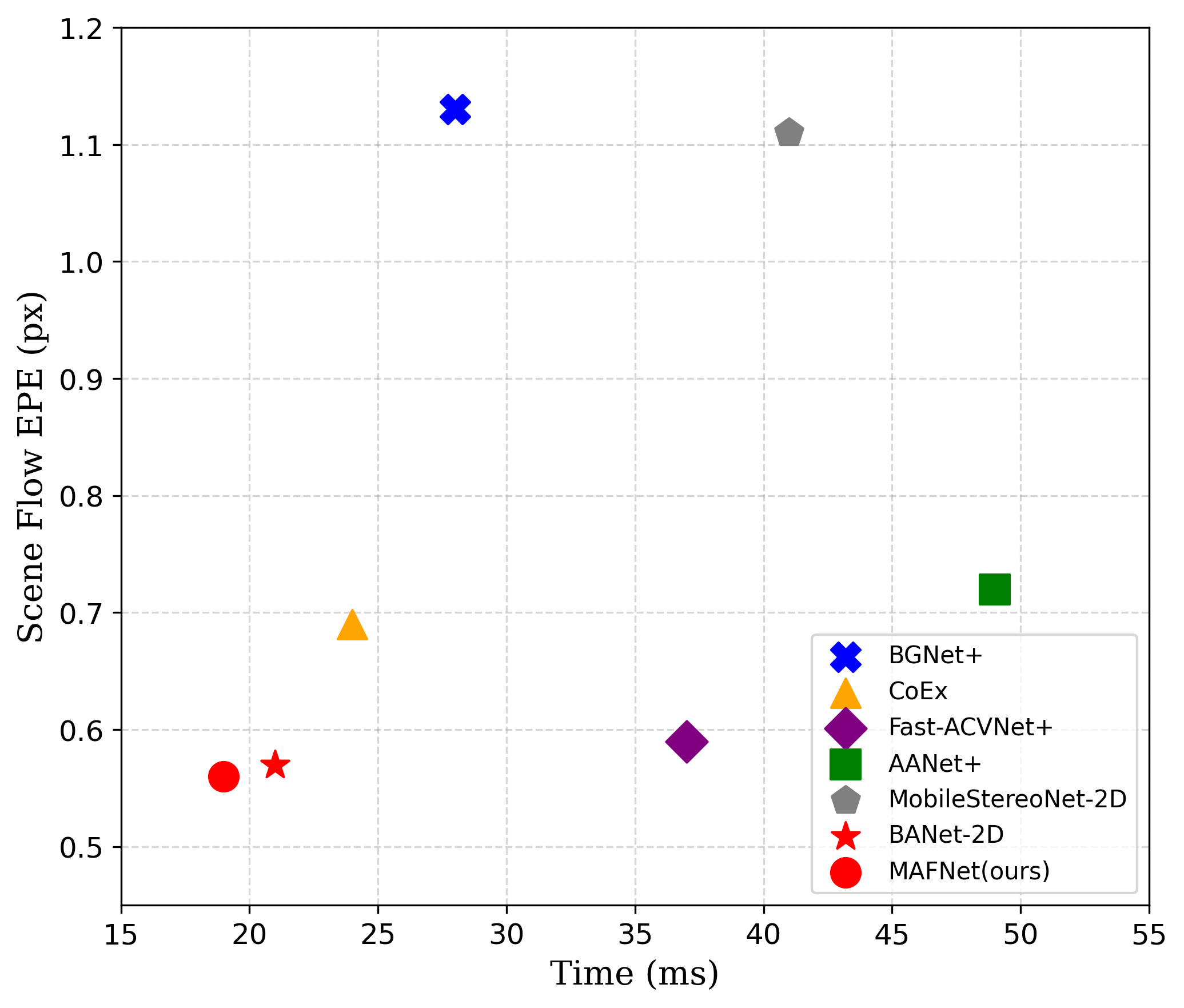}
\caption{Comparison with state-of-the-art real-time methods on the Scene Flow (4090GPU).}
\label{fig:sf_com}
\end{figure}

To improve efficiency, subsequent studies have introduced lightweight strategies such as constructing lower-resolution cost volumes, adopting sparse disparity searches, and employing lightweight aggregation networks like StereoNet \cite{khamis2018stereonet}, DeepPruner \cite{duggal2019deeppruner}, and Fast-ACVNet \cite{xu2023accurate}. While these methods achieve faster inference on high-end GPUs, they still depend on extensive 3D convolutions or complex deformable operations, limiting deployment on mobile platforms, as shown in Fig. 1. Another line of work replaces 3D with 2D convolutions for cost aggregation, e.g., AANet \cite{xu2020aanet} uses deformable convolutions for adaptive aggregation, and MobileStereoNet-2D \cite{shamsafar2022mobilestereonet} integrates lightweight MobileNet modules. Although these approaches improve efficiency, they remain prone to errors in blurred textures, textureless regions, and fine edges, leading to noticeable detail loss.

To overcome these limitations, MAFNet introduces an adaptive frequency-domain attention to explicitly separate high-frequency details from low-frequency regions and a Linformer-based fusion module for lightweight, dynamic aggregation. This design retains the efficiency of 2D convolutions while strengthening the modeling of complex textures and edges, thus achieving a better trade-off between real-time performance and accuracy and making the network well suited for edge or mobile deployment. Our contributions are as follows:
\begin{itemize}
\item We propose MAFNet, a lightweight stereo matching framework that achieves accurate disparity estimation using only 2D convolutions, enabling efficient deployment on mobile and edge devices.

\item An Adaptive Frequency-Domain Filtering Attention (AFFA) module is introduced to separate high-frequency details from low-frequency smooth regions, reducing errors in textureless areas and near thin edges.

\item An Adaptive Aggregation of High- and Low-Frequency (AAHF) module is designed based on a Linformer architecture to enable global feature interaction with low computational cost, achieving a better balance between runtime efficiency and prediction accuracy compared with existing mobile-oriented methods.
\end{itemize}

\section{Method}

\subsection{Model Architecture}

\begin{figure*}[t]
    \centering
    \includegraphics[width=\textwidth]{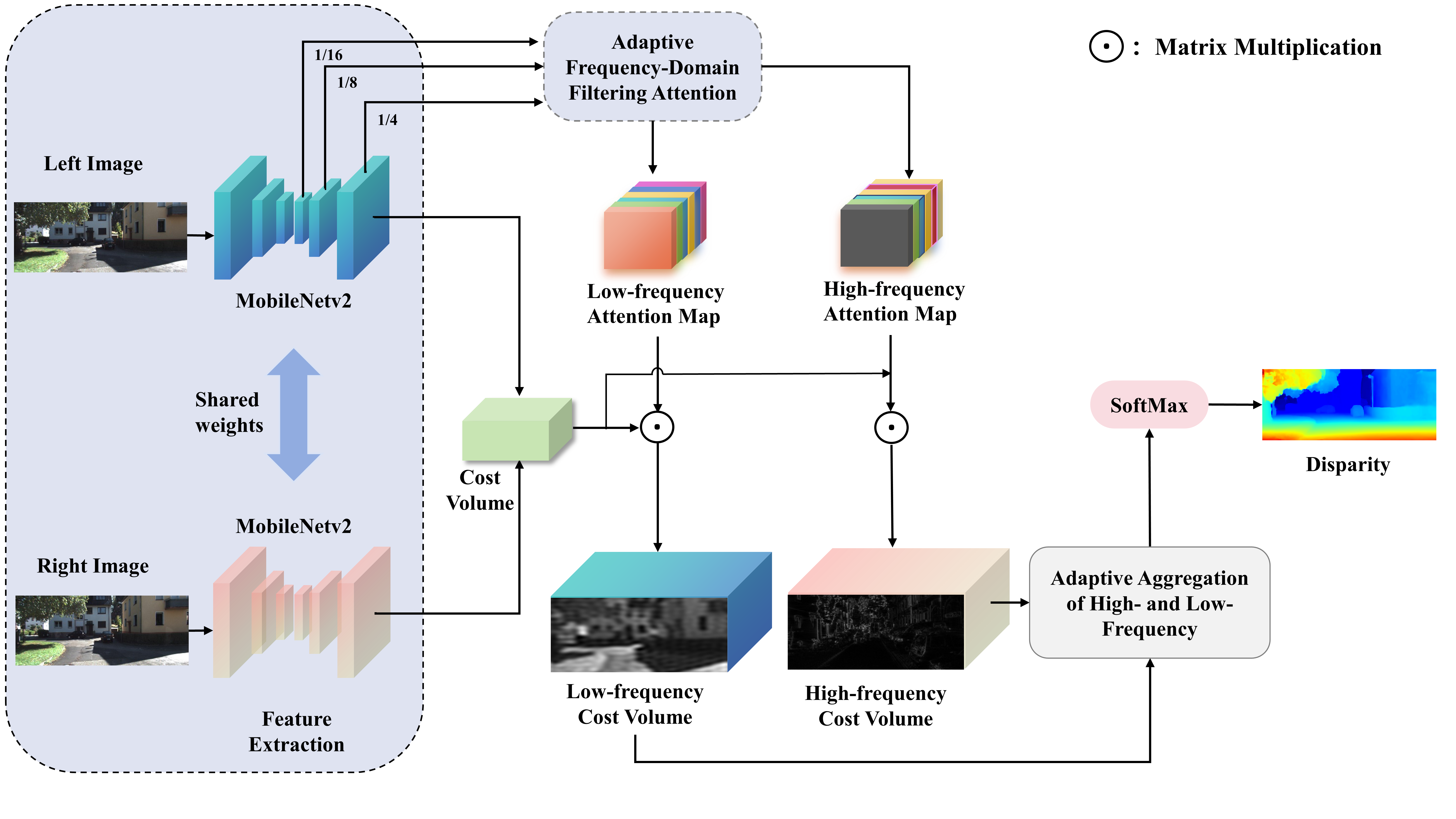}
    \caption{Overview of the proposed network. Given a rectified stereo image pair, multi-scale features are extracted using a weight-sharing MobileNetv2 backbone. An adaptive frequency-domain attention separates features into high- and low-frequency components, which are used to construct corresponding cost volumes. These cost volumes are aggregated by a high–low frequency adaptive fusion module, followed by disparity regression to generate the final disparity map.}
    \label{fig:overall}
\end{figure*}

\textbf{Feature Extraction.} As shown in Fig. 2, our network takes a pair of rectified images as input and first extracts multi-scale high-dimensional features using a weight-sharing MobileNetv2 backbone. Specifically, feature maps are obtained at four scales: $1/4$, $1/8$, $1/16$, and $1/32$ of the original resolution. Starting from the 1/32-resolution features, we progressively upsample to 1/4 resolution via transposed convolutions with a 4×4 kernel and stride 2, producing multi-scale features {$f_l$, $f_r$} for the left and right images, with three key scales ($1/4$, $1/8$, $1/16$) retained. The multi-scale features of the left image are fed into an adaptive frequency-domain filtering attention module to generate frequency-aware spatial weights, while the 1/4-resolution features of both images are used to construct the cost volume.

\textbf{Cost Volume Construction.} Given the quarter-resolution feature maps of the left and right images, $f_{l,4}, f_{r,4} \in \mathbb{R}^{B \times N_c \times H/4 \times W/4}$, we construct the group-wise correlation volume by first evenly dividing the feature channels into $N_g$ groups, each containing $N_c / N_g$ channels, which is defined as:
\begin{equation}
C(x, y, d, g)
=
\frac{1}{N_c / N_g}
\left\langle
f_{l,4}^{(g)}(x, y),
f_{r,4}^{(g)}(x - d, y)
\right\rangle ,
\end{equation}
where $(x,y)$ denotes the spatial location, $\langle \cdot, \cdot \rangle$ denotes the inner product, $d \in \mathcal{D} = \{0, 1, 2, \ldots, D/4 - 1\}$ is the disparity index, $g \in \{1, 2, \ldots, N_g\}$ denotes the group index.

\textbf{Disparity Prediction.} The proposed AFFA is used to construct the high- and low-frequency cost volumes. The aggregated cost volume is then obtained through the proposed AAHF module, followed by a softmax operation for disparity regression to produce the final disparity map $d_{0}$. The architectural details of the AFFA and AAHF modules are presented in Sec. 2.2 and 2.3, respectively:

\begin{equation}
D_0 = \sum_{d=0}^{D/4-1} d \times \mathrm{Softmax}\!\left(C_{\mathrm{AAHF}}(d)\right),
\label{eq:disp_regression}
\end{equation}
where $D$ denotes the predefined maximum disparity, and $d$ indexes the disparity candidates within the range $[0, D/4 - 1]$. The predicted disparity map $D_0$ is obtained at a resolution of $H/4 \times W/4$. To recover a full-resolution result, superpixel-based weights are applied around each pixel in the left image to aggregate local neighboring values in $D_0$, yielding the final disparity map $D_1 \in \mathbb{R}^{H \times W}$.

\textbf{Loss Function.} The entire network is trained in a supervised end-to-end manner, and the final loss function is defined as:
\begin{equation}
L=\ \lambda_0L_{1\ smooth}(D_0-D_{gt}^{1/4})+\lambda_1L_{1\ smooth}(D_1-D_{gt})
\label{eq:loss}
\end{equation}
where $D_{gt}$ denotes the ground-truth disparity map and $L_{1\text{-smooth}}$ is the smooth L1 loss. $D_{gt}^{1/4}$ represents the ground-truth disparity downsampled to $1/4$ resolution. $D_0$ and $D_1$ denote the predicted disparities at $1/4$ and full resolution, respectively.

\begin{table*}[t]
\centering
\caption{Results of ablation experiment on the Scene Flow dataset.}
\label{tab:ablation}
\renewcommand{\arraystretch}{1.2}
\setlength{\tabcolsep}{8pt}
\begin{tabular}{c c c c c c}
\toprule
\textbf{Model} &
\textbf{Adaptive Frequency-Domain} &
\textbf{Adaptive Aggregation} &
\textbf{EPE (px)} &
\textbf{Bad 3.0 (\%)} &
\textbf{Param(M)} \\
&
\textbf{Filtering Attention} &
\textbf{of High- and Low-Frequency} &
&
&
\\
\midrule
Baseline          & $\times$      & $\times$      & 0.64 & 2.87 & 3.69 \\
AFFA                & \checkmark    & $\times$      & 0.58 & 2.53 & 3.81 \\
AFFA+AAHF (MAFNet) & \checkmark    & \checkmark    & \textbf{0.56} & \textbf{2.41} & 3.82 \\
\bottomrule
\end{tabular}
\end{table*}

\begin{table}[t]
\centering
\caption{Results of ablation experiment on KITTI 2012 and  KITTI 2015 datasets.}
\label{tab:kitti_single}
\renewcommand{\arraystretch}{1.2}
\setlength{\tabcolsep}{6pt}
\begin{tabular}{c cc ccc}
\toprule
\multirow{2}{*}{\textbf{Method}} &
\multicolumn{2}{c}{\textbf{KITTI 2012}} &
\multicolumn{3}{c}{\textbf{KITTI 2015}} \\
& \textbf{3-noc} & \textbf{3-all} & \textbf{D1-bg} & \textbf{D1-fg} & \textbf{D1-all} \\
\midrule
w/o AFFA    & 1.73 & 2.18 & 1.81 & 3.75 & 2.13 \\
MAFNet  & \textbf{1.36} & \textbf{1.76} & \textbf{1.57} & \textbf{3.02 (19\%$\uparrow$)} & \textbf{1.81} \\
\bottomrule
\end{tabular}
\end{table}

\subsection{Adaptive Frequency-Domain Filtering Attention}

High-frequency image components mainly encode fine details and edges, while low-frequency components capture global structures and textureless regions. To leverage these complementary features, we propose an adaptive frequency-domain filtering attention module. The module first decomposes feature maps in the frequency domain to obtain high- and low-frequency subbands. A learnable spatial gating mechanism is then applied to dynamically generate pixel-wise fusion weights for the two subbands. The resulting attention maps adaptively distinguish frequency characteristics across different regions, effectively enhancing edge fidelity and improving matching stability in textureless areas.

\begin{figure}[htbp]
\centering
\includegraphics[width=1\columnwidth]{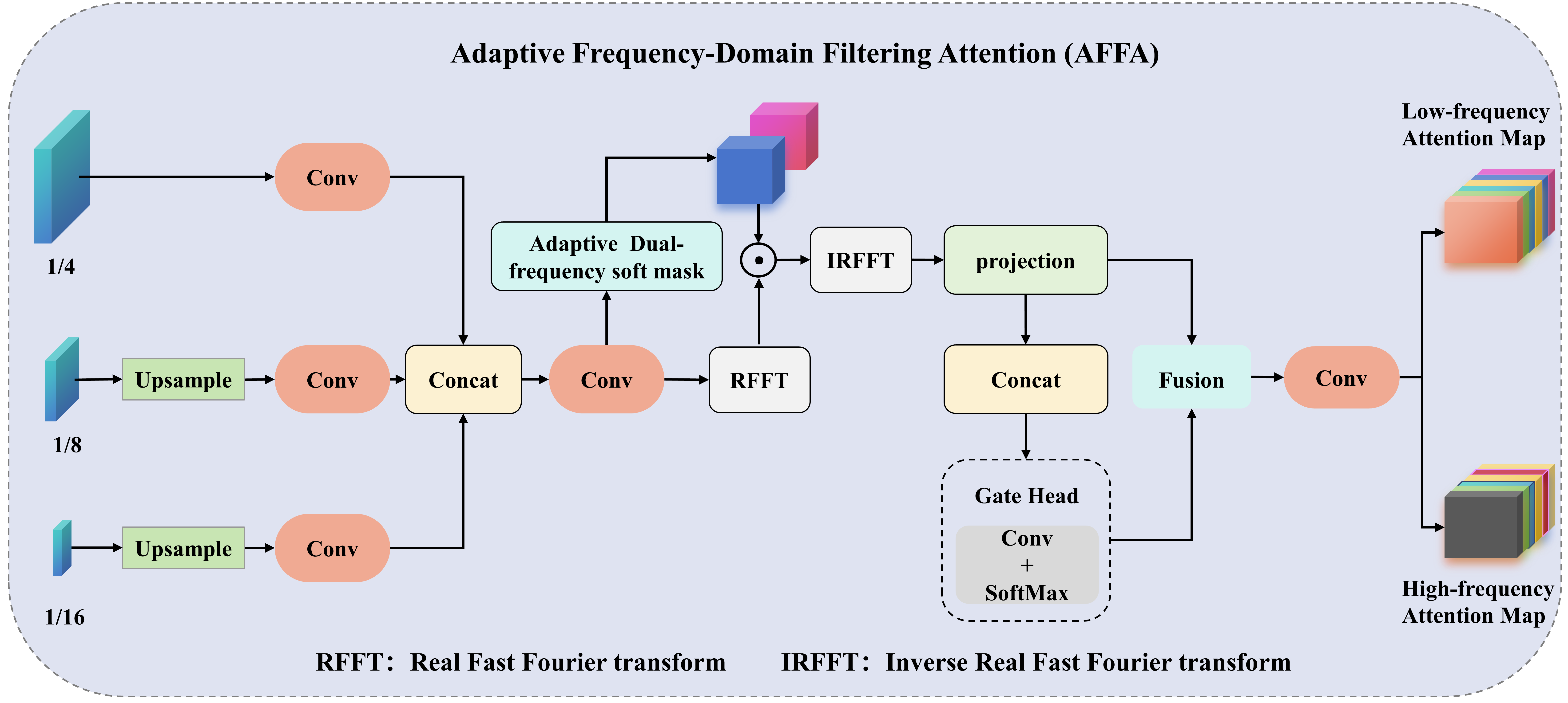}
\caption{AFFA Structure Diagram.}
\label{fig:AFFA}
\end{figure}

As illustrated in Fig. 3, given a feature map $X\in\mathbb{R}^{B \times C \times H \times W}$, we first compute its 2D Real Fast Fourier Transform (RFFT) to obtain the frequency-domain representation:
\begin{equation}
F(X)=RFFT{\left(X\right),F}(X)\in\mathbb{C}^{B\times C\times H\times W}
\label{eq:RFFT}
\end{equation}
In the frequency plane, the low-frequency components are concentrated near the center while the high-frequency components are distributed toward the edges. We introduce learnable thresholds $\tau_{l}$ and $\tau_{h}$ to adaptively separate the bands and construct soft masks as:
\begin{equation}
M_{low}(u,v)=\ \sigma(\frac{\tau_{l}-r(u,v)}{\gamma})
\label{eq:mlow}
\end{equation}
\begin{equation}
M_{high}(u,v)=\sigma(\frac{r(u,v)-\tau_h}{\gamma})
\label{eq:mhigh}
\end{equation}
where $r(u,v)$ denotes the normalized radius from a frequency-domain point to the center, $\sigma$ is the sigmoid function, and $\gamma$ is the temperature parameter used to control the smoothness of the soft threshold. Thus, the high- and low-frequency decomposition is obtained:
\begin{equation}
X_{low}=IRFFT{(F\left(X\right)}\odot M_{low})
\label{eq:xlow}
\end{equation}
\begin{equation}
X_{high}=IRFFT{(F\left(X\right)}\odot M_{high})
\label{eq:xhigh}
\end{equation}
where IRFFT denotes the inverse real fast Fourier transform. To avoid the constraints introduced by fixed frequency band division, we propose a spatially adaptive gating mechanism. Specifically, the high- and low-frequency features are first concatenated:
\begin{equation}
Z = Concat \!\left[X_{low},\, X_{high}\right] \in \mathbb{R}^{B \times 2C \times H \times W}
\label{eq:z}
\end{equation}
Then, pixel-level weights are obtained via a $1\times1$ convolution:
\begin{equation}
G = Softmax \!\left(Conv_{1\times 1}(Z)\right) \in \mathbb{R}^{B \times 2 \times H \times W}
\label{eq:G}
\end{equation}
\begin{equation}
G_{\text{low}} = G[:,\,0,\,:,:], \quad
G_{\text{high}} = G[:,\,1,\,:,:]
\label{eq:glowhigh}
\end{equation}
The fused feature is given by:
\begin{equation}
X_f=G_{low}\odot X_{low}+G_{high}\odot X_{high}
\label{eq:xf}
\end{equation}
Finally, convolutional layers followed by a sigmoid function are applied to generate the high- and low-frequency attention maps:
\begin{equation}
A_{high}=\sigma\left(Conv\left(X_f\right)\right)
\label{eq:ahigh}
\end{equation}
\begin{equation}
A_{low}=1-A_{high}
\label{eq:alow}
\end{equation}

\subsection{Adaptive Aggregation of High- and Low-Frequency}

\begin{figure}[htbp]
\centering
\includegraphics[width=1\columnwidth]{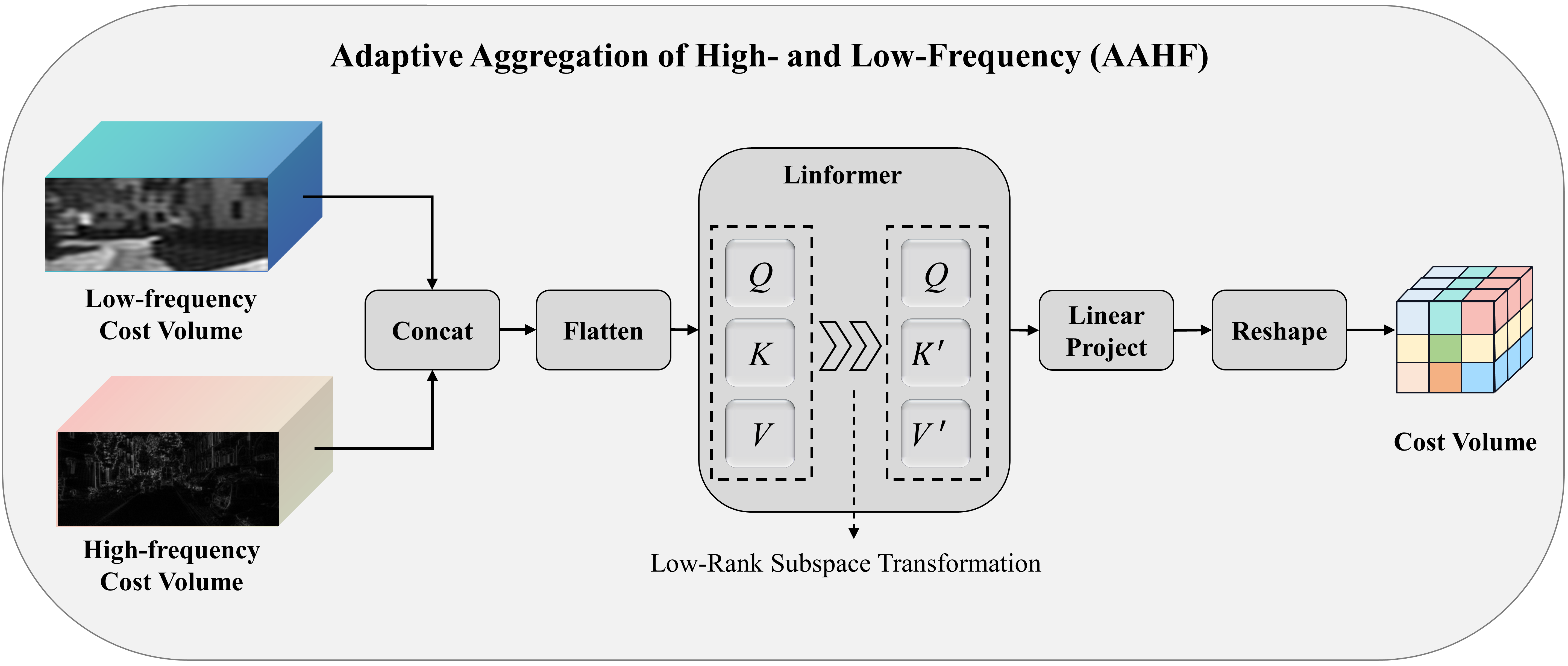}
\caption{AAHF Structure Diagram.}
\label{fig:AAHF}
\end{figure}

\begin{table*}[t]
\centering
\caption{Quantitative comparison on the KITTI 2012 and KITTI 2015 benchmarks.}
\label{tab:kitti_results}

\setlength{\tabcolsep}{5.5pt}       
\renewcommand{\arraystretch}{1.25}   

\resizebox{\textwidth}{!}{
\begin{tabular}{l|cccccc|ccc|c}
\hline
Method 
& \multicolumn{6}{c|}{KITTI 2012 \cite{geiger2012we}} 
& \multicolumn{3}{c|}{KITTI 2015 \cite{menze2015object}} 
& Time \\
& 3-noc & 3-all & 4-noc & 4-all 
& Avg-Noc & Avg-All 
& D1-bg & D1-fg & D1-all & (ms)\\
\hline
CoEx \cite{bangunharcana2021correlate}
& 1.55 & 1.93 & 1.15 & 1.42 & 0.5 & 0.5 & 1.79  & 3.82  & 2.13  & 24\\

AANet+ \cite{xu2020aanet}
& 1.55 & 2.04 & 1.20 & 1.58 & 0.4 & 0.5 & 1.65 & 3.96 & 2.03 & 49 \\

DeepPruner-Fast \cite{duggal2019deeppruner}
& -- & -- & -- & -- & -- & -- & 2.32 & 3.91	& 2.59 & 43 \\

Fast-ACVNet \cite{xu2022attention}
&1.68 &2.13 &1.23 &1.56 &0.5 &0.6 &1.82 &3.93 &2.17 & 31 \\

Fast-ACVNet+ \cite{xu2022attention}
&1.45 &1.85 &1.06 &1.36 &0.5 &0.5 &1.70 &3.53 &2.01 & 37 \\

HITNet \cite{tankovich2021hitnet}
&1.41 &1.89 &1.14 &1.53 &0.4 &0.5 &1.74 &3.20 &1.98 & \textbf{12} \\

MobileStereoNet-2D \cite{shamsafar2022mobilestereonet}
& -- & -- & -- & -- & -- & -- &2.49 &4.53 &2.83    & 41   \\

BGNet+ \cite{xu2021bilateral}
&1.62 &2.03 &1.16 &1.48 &0.5 &0.6 &1.81 &4.09 &2.19 & 28 \\

BANet-2D \cite{xu2025banet}
&1.38 &1.79 &1.01 &1.32 &0.5 &0.5 &1.59 &3.03 &1.83 & 21 \\

\hline
MAFNet (Ours) 
& \textbf{1.36} & \textbf{1.76} & \textbf{0.98} & \textbf{1.31} 
& 0.5 & 0.5 
& \textbf{1.57} & \textbf{3.02} & \textbf{1.81} 
& 19 \\
\hline
\end{tabular}
}
\end{table*}

\begin{figure*}[htbp]
    \centering
    \includegraphics[width=\textwidth]{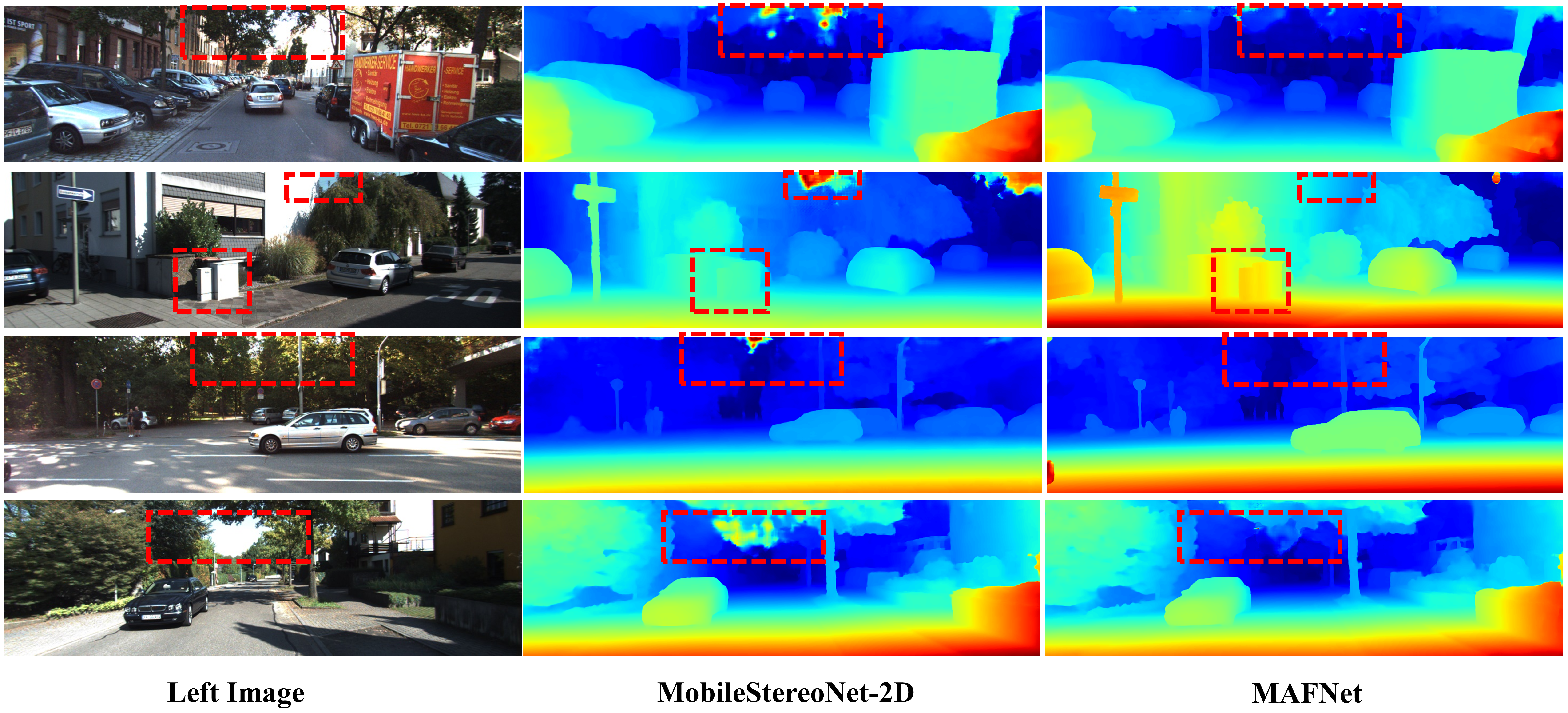}
    \caption{Qualitative evaluation results on the KITTI 2012 \cite{geiger2012we} and KITTI 2015 \cite{menze2015object} benchmarks. Through the introduction of a high- and low-frequency separation mechanism, MAFNet produces well-defined edges while maintaining complex structural details, enabling reliable matching even in extensive textureless regions and under drastic lighting changes.}
    \label{fig:results}
\end{figure*}

The high-frequency feature and low-frequency feature respectively capture detailed edges and smooth regions, and they are complementary in stereo matching. Simply using element-wise addition or multiplication for fusion often fails to adaptively adjust their contributions across different regions, resulting in insufficient feature aggregation. To address this, we introduce a Linformer-based adaptive fusion module, which employs low-rank attention mapping to achieve efficient and dynamic feature fusion.

Given the constructed raw cost volume $C$, it is decomposed into a high-frequency cost volume ${CV}_{high}$ and a low-frequency cost volume ${CV}_{low}$:
\begin{equation}
\begin{aligned}
{CV}_{high} &= A_{high} \odot C, \\
{CV}_{low} &= A_{low} \odot C
\end{aligned}
\label{eq:clowhigh}
\end{equation}
As shown in Fig. 4, given the input high-frequency cost volume ${CV}_{high}\in\mathbb{R}^{B \times D\times H\times W}$ and low-frequency cost volume ${CV}_{low}\in \mathbb{R}^{B \times D\times H\times W}$, we first concatenate them to obtain:
\begin{equation}
{CV}_{all} = Concat({CV}_{high},\ {CV}_{low})\ \in\mathbb{R}^{B\times2D \times H \times W}
\label{eq:cvall}
\end{equation}
It is then flattened into a sequential representation:
\begin{equation}
X = \mathrm{Reshape}\!\left(\mathrm{CV}_{\mathrm{all}}\right) \in \mathbb{R}^{B \times N \times 2D}, \quad N = H \times W 
\label{eq:flatten}
\end{equation}
The standard self-attention is defined as:
\begin{equation}
Att(Q,K,V)=SoftMax(\frac{QK^T}{\sqrt d})V
\label{eq:att}
\end{equation}
where $Q,\ K,\ V\in\mathbb{R}^{N \times d}$, with a computational complexity of $O(N^2)$. In Linformer, the key $K$ and value $V$ are projected into a low-rank subspace:
\begin{equation}
K' = E K, \quad V' = F V, \quad (E, F) \in \mathbb{R}^{k \times N}, \quad k \ll N
\label{eq:kv}
\end{equation}
The attention computation is given by:
\begin{equation}
LinAtt(Q, K, V)
= Softmax\!\left(\frac{Q {K'}^{T}}{\sqrt{d}}\right) V'
\label{eq:lin}
\end{equation}
The complexity is reduced to $O(N \cdot k)$. Through the low-rank approximation, the model can significantly reduce computational cost while maintaining representational capacity. Finally, we use a linear projection to map the output back to $\mathbb{R}^{B\times N\times D}$ and reshape it into the fused volume $C_{AAHF}$:
\begin{equation}
C_{AAHF} = Reshape\!\left(LinAtt(X)\right) \in \mathbb{R}^{B \times D \times H \times W}
\label{eq:fused}
\end{equation}

\section{Experiments}

\subsection{Implementation Details and Datasets}
The proposed MAFNet model was implemented using PyTorch, and experiments were carried out on the Scene Flow \cite{mayer2016large}, KITTI 2012 \cite{geiger2012we}, and KITTI 2015 \cite{menze2015object} datasets. All experiments were conducted on an NVIDIA RTX 4090 GPU. We employed the AdamW \cite{loshchilov2017decoupled} optimizer combined with a one-cycle learning rate scheduler with warm-up, setting the maximum learning rate to 8e-4. The weights in the loss function were configured as $\lambda_0\ =0.3$ and $\lambda_1\ =1.0$. During training, we initially pre-trained on the Scene Flow dataset for 200k steps with a batch size of 16 to adequately learn the distribution of large-scale synthetic data. The resulting model was then fine-tuned for 50k steps on the mixed KITTI 2012 and KITTI 2015 training sets to improve performance in real-world environments. During training, each input image was randomly cropped to a resolution of 256×512.

\subsection{Ablation Study}
Comprehensive ablation studies were carried out on Scene Flow \cite{mayer2016large} and KITTI \cite{geiger2012we, menze2015object} to verify the effectiveness of the proposed approach, with the ablation results for Scene Flow presented in Table I and those for KITTI 2012 \cite{geiger2012we} and 2015 \cite{menze2015object} presented in Table II. The baseline adopts a single-branch aggregation architecture.  Without AFFA, the input to AAHF is a standard cost volume.

Table I presents the ablation study results evaluated on the Scene Flow dataset. The AFFA module decouples high-frequency details and low-frequency smooth regions, leading to a decrease in EPE from 0.64 to 0.58 and Bad 3.0 error from 2.87\% to 2.53\% while introducing only a marginal number of additional parameters. By further fusing multi-frequency information, the AAHF module lowers the EPE to 0.56 and the Bad 3.0 error to 2.41\%, incurring an almost negligible increase in model parameters.

Table II reports the results on the KITTI 2012 \cite{geiger2012we} and KITTI 2015 \cite{menze2015object} datasets. These datasets reflect real driving scenes. The performance gains are larger than those on Scene Flow \cite{mayer2016large}. On KITTI 2015 \cite{menze2015object}, the D1-fg metric evaluates errors in foreground regions. These regions contain more object boundaries and fine details. The D1-fg value drops from 3.75\% to 3.02\%. This result corresponds to a 19\% improvement. The improvement shows that the proposed method handles high-frequency details and object edges more effectively in foreground areas.

\subsection{Comparisons with State-of-the-Art Methods}
\textbf{Quantitative Comparisons.} Table ~\ref{tab:kitti_results} presents quantitative comparisons on the KITTI 2012 \cite{geiger2012we} and KITTI 2015 \cite{menze2015object} benchmarks. MAFNet achieves the best overall performance among real-time and mobile-oriented stereo methods, while maintaining efficient inference speed. In particular, on KITTI 2015 \cite{menze2015object}, MAFNet obtains the lowest D1-all error of 1.81\% with only 19 ms runtime, surpassing MobileStereoNet-2D \cite{shamsafar2022mobilestereonet} by 36.0\% and BANet-2D \cite{xu2025banet} by 1.1\%. These results demonstrate that MAFNet achieves a superior accuracy–efficiency trade-off and is well suited for real-time deployment on mobile and edge platforms.  As illustrated in Fig. 1, the comparison results on Scene Flow demonstrate the effectiveness of the proposed approach.

\textbf{Qualitative Comparisons.} The visual performance of the proposed MAFNet is compared with that of the mobile-oriented MobileStereoNet-2D \cite{shamsafar2022mobilestereonet}. As illustrated in Fig. 5, owing to its single feature aggregation mechanism, MobileStereoNet-2D tends to produce edge blurring and structural detail loss in high-frequency edges, thin structures, and large textureless areas. By contrast, thanks to the explicit decoupling and adaptive fusion of high- and low-frequency components, MAFNet can more precisely capture object boundaries while retaining intricate structural details. It exhibits more stable and coherent disparity estimation in high-frequency regions such as tree branches and traffic signs, as well as in scenes with significant illumination variations, resulting in more reliable matching performance.

\section{Conclusion}
In this work, we introduced MAFNet, a novel stereo matching network that effectively balances accuracy and efficiency for real-time applications on resource-constrained devices. By leveraging an adaptive frequency-domain filtering attention module and a Linformer-based low-rank fusion mechanism, MAFNet can separately process high- and low-frequency information and integrate them for robust disparity estimation, while avoiding the computational burden of 3D convolutions. Extensive experiments on Scene Flow and KITTI 2015 benchmarks validate that MAFNet consistently outperforms existing real-time stereo matching methods, achieving superior accuracy with reduced computational cost. This demonstrates the potential of adaptive frequency-domain filtering attention and adaptive aggregation of high- and low-frequency in designing lightweight yet high-performance stereo networks, providing a promising solution for real-world mobile and embedded vision applications.

\bibliographystyle{IEEEbib}
\bibliography{icme2026references}

\end{document}